# Enhancing Scalability and Reliability in Semi-Decentralized Federated Learning With Blockchain: Trust Penalization and Asynchronous Functionality

Ajay Kumar Shrestha
*Computer Science Department*
*Vancouver Island University*
Nanaimo, Canada
ajay.shrestha@viu.ca

Faijan Ahamad Khan
*Bayes Solutions*
Los Angeles, USA
faijian@bayes.global

Mohammed Afaan Shaikh
*Bayes Solutions*
Los Angeles, USA
afaan@bayes.global

Amir Jaberzadeh
Bayes Solutions
Los Angeles, USA
amir@bayes.global

Jason Geng
Bayes Solutions
Los Angeles, USA
jason@bayes.global

*Abstract*—The paper presents an innovative approach to address the challenges of scalability and reliability in Distributed Federated Learning by leveraging the integration of blockchain technology. The paper focuses on enhancing the trustworthiness of participating nodes through a trust penalization mechanism while also enabling asynchronous functionality for efficient and robust model updates. By combining Semi-Decentralized Federated Learning with Blockchain (SDFL-B), the proposed system aims to create a fair, secure and transparent environment for collaborative machine learning without compromising data privacy. The research presents a comprehensive system architecture, methodologies, experimental results, and discussions that demonstrate the advantages of this novel approach in fostering scalable and reliable SDFL-B systems.

*Keywords—blockchain, smart contracts, machine learning, distributed federated learning, trust, incentives*

## I. Introduction

In recent years, decentralized approaches have gained significant attention in the field of machine learning. The traditional centralized paradigm faces challenges in handling extensive datasets, addressing data privacy concerns, and relying on a single central authority [1]. Decentralized machine learning, particularly Semi-Decentralized Federated Learning (SDFL), offers a promising solution to these issues by distributing model training across multiple devices, allowing data to remain on-device and preserving privacy while enabling collaborative learning [2], [3]. The SDFL is a strategic amalgamation of decentralization and controlled coordination allowing local model training and parameter exchange while designating an aggregator for model aggregation [3]. The Federated Learning algorithm preserves user privacy by avoiding raw data collection, but recent studies highlight potential vulnerabilities in parameter-based attacks, underscoring the need for further advancements in federated learning frameworks [4].

The combination of Semi-Decentralized Federated Learning with Blockchain (SDFL-B) holds great potential for overcoming existing challenges in Distributed Federated Learning (DFL) [5]. Leveraging blockchain's attributes of immutability and transparency, this integration establishes a secure and tamper-resistant platform for cultivating trust and encouraging honest participation in the decentralized learning process [6]. By incorporating blockchain technology into SDFL, we seek to address the issues of reliability and trust, promoting fairness and accountability among participating nodes.

This paper sets out to explore and propose an innovative approach addressing two critical aspects of SDFL: scalability and reliability. We investigate the challenges associated with scalability in SDFL and present a robust architecture that can efficiently manage an increasing number of participating nodes. The proposed system leverages distributed computation and communication techniques to ensure seamless scaling for large-scale SDFL deployments. To enhance the trustworthiness of nodes participating in the SDFL process, we introduce a trust penalization mechanism. This mechanism identifies and penalizes untrustworthy nodes based on their contributions, discouraging dishonest behavior and promoting a reliable and cooperative learning environment. To further enhance the robustness and real-time performance of SDFL, we incorporate asynchronous functionality. This allows nodes to contribute model updates at their own pace, making the system resilient against node failures and network delays. Existing literature has yet to cover a standard federation approach for various machine learning (ML) frameworks, especially in the context of DFL [3]. Therefore, there is a need to create and implement a robust and adaptive codebase for producing generic ML scenarios. Our goal is to develop an adaptable and expandable solution applicable to different frameworks and application scenarios, advancing decentralized machine-learning techniques. This will lead to scalable and reliable SDFL systems capable of accommodating diverse applications across various domains.

The rest of the paper is organized as follows. Section II provides a succinct analysis of existing architectures and identifies their shortcomings, laying the groundwork for the need of a novel solution. Our proposed model for the solution architecture is detailed in Section III, outlining the integration of SDFL with blockchain to address the identified challenges. This section also explains the trust penalization mechanism, illustrating how it fosters reliability and accountability among participating nodes, as well as the implementation of asynchronous functionality, showcasing its benefits in achieving real-time performance and robustness. The experimental setup is presented in Section IV, while Section V offers results and analysis, validating the effectiveness of our approach in promoting the scalability and trustworthiness of SDFL-B. Section VI discusses the findings, and lastly, Section VII concludes the paper by summarizing key contributions and outlining potential future research directions in this dynamic field.

## II. BACKGROUND AND RELATED WORKS

Distributed Federated Learning (DFL), initially introduced in 2018, aimed to decentralize the aggregation of model parameters among neighboring participants [7]. In DFL, the core operation involves the rapid transmission of locally computed updates from each node, such as model parameters or gradients, and accompanying metadata, like activation functions in neural networks, to other federation nodes. In comparison to Centralized Federated Learning (CFL), DFL effectively tackles issues associated with single points of failure, trust dependencies, and bottlenecks that can occur at the server node [8].

DFL brings improvements in fault tolerance by enabling nodes to maintain an updated awareness of available or inactive communicating nodes [9], thus reducing vulnerability to single-point attacks. Additionally, DFL mitigates network bottleneck challenges by evenly distributing communication and workloads across nodes, thereby minimizing the risk of congestion or performance delays across the network [10].

However, alongside these advantages, DFL introduces novel challenges, including increased communication overhead, the optimization of training processes, and the assurance of trustworthy AI. Depending on how model aggregation is distributed within the network, specific DFL configurations may experience elevated communication overhead [11]. In such scenarios, careful planning, fine-tuning of communication protocols, client selection strategies, and trust mechanisms become crucial for mitigating these limitations. DFL encompasses various dimensions, including network topology defining node associations [12], communication mechanisms coordinating model parameter exchange [1], and security and privacy, covering potential cyberattacks and measures to safeguard data privacy and model robustness [13].

Within this framework, three key perspectives to be explored: nodes, communications, and models. The first perspective involves assessing the diversity and dynamism of nodes within DFL. The second focuses on the effectiveness of inter-node communications for data exchange. The third centers on evaluating the performance of machine learning and deep learning models in collaborative task-solving.

In a prior study [14], the authors employed a semi-decentralized federated learning algorithm where clients collaborated by relaying neighboring updates to a central parameter server (PS), intending to mitigate the impact of intermittent connectivity issues and improve convergence rates. Clients computed local consensus from neighbors' updates and sent a weighted average of their own and neighbors' updates to the PS. The algorithm was optimized for these averaging weights to achieve unbiased global updates, enhancing convergence rates and reducing variance.

Another paper [15] introduced the concept of Federated Learning Empowered Overlapped Clustering for Decentralized Aggregation (FL-EOCD). This approach leveraged device-to-device (D2D) communications and overlapped clustering to achieve decentralized aggregation, eliminating the need for a central aggregator. The paper also presented an energy-efficient framework for FL-EOCD within a partially connected D2D network, with a specific focus on addressing energy consumption and convergence rate.

In [16], a semi-decentralized learning method merging device-to-server and device-to-device (D2D) communication for model training was introduced. This approach involved local training and D2D-based consensus among device clusters, addressing issues related to diverse resources and device proximity. It demonstrated improved model accuracy, training time, and energy efficiency compared to existing methods. However, limitations included the need to address scalability challenges, potential overhead from D2D communications, and the robustness of the approach in dynamic or adversarial scenarios.

In a different study [17], an asynchronous federated learning aggregation protocol utilizing a permissioned blockchain was introduced. This protocol integrated the learned model into the blockchain and performed two-order aggregation calculations, effectively reducing synchronization problems. However, addressing challenges related to scalability, diverse network conditions, and various data types, as well as optimizing system performance, remains vital for future research in real-world edge computing scenarios.

## III. SYSTEM MODEL

### A. System Architecture

In this paper, we introduce the blockchain-coupled cluster-based semi-decentralized federated learning architecture, a novel approach that capitalizes on geographical proximity to optimize the efficiency and communication dynamics of the federated learning process.

As shown in Fig. 1, clusters of participating workers are formed based on their geographical locations. These clusters create localized groups that foster efficient resource allocation and minimize communication overhead. Within each cluster, a randomly designated worker is assigned the role of cluster head. The cluster head takes charge of coordinating the federated learning process within their respective cluster. Central to this role is the aggregation of model weights contributed by individual workers during their local training processes. After collecting these updated model weights, the cluster head initiates the aggregation process. The aggregated model is then disseminated not only among the cluster's workers but also stored externally on the InterPlanetary File System (IPFS) platform. This ensures accessibility to the aggregated model for all cluster workers, even in the presence of intermittent network disruptions.

Importantly, communication extends beyond individual clusters. Workers from one cluster can request the model's hash from another cluster's head if they find value in the aggregated model generated by the latter. Upon obtaining the hash, the requesting cluster's head incorporates the model into its own aggregation process, enhancing the collaborative essence of the semi-decentralized federated learning.

To maintain dynamism and diversity, the worker head selection process is cyclic. The current cluster head periodically reshuffles and designates a new worker head from the participant pool. This rotation ensures that the model aggregation process remains dynamic and equitable, preventing any single worker from exerting consistent influence over the model.

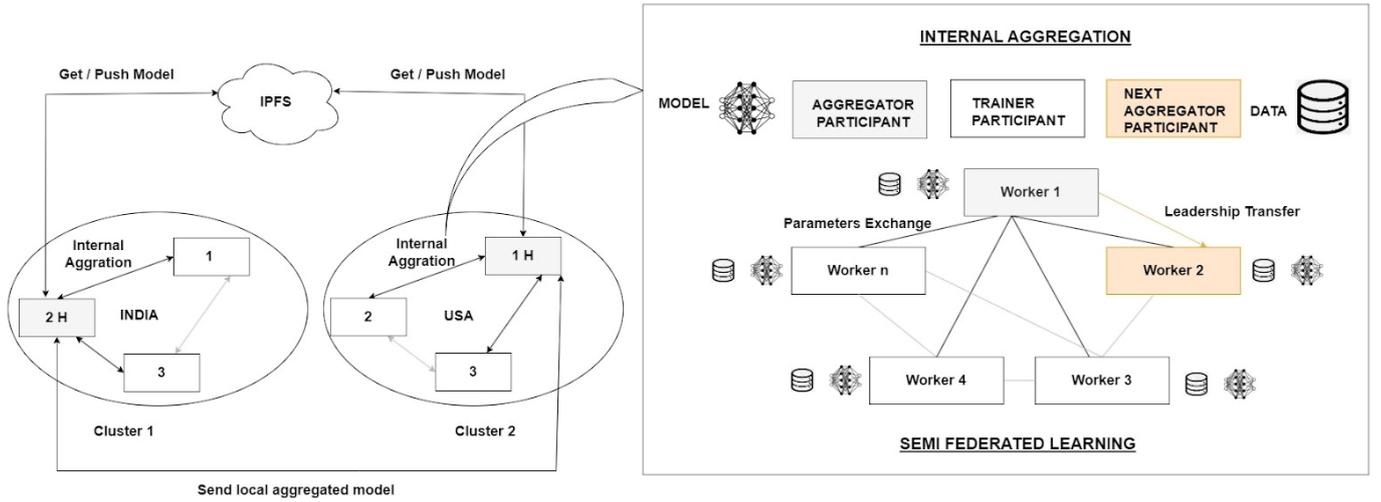

Fig. 1. Network Architecture — Cluster-Based Semi-Decentralized Federated Learning

*B. System Workflow*

In this framework, multiple workers establish peer-to-peer connections bypassing the need for an intermediary or central server. The federated learning process begins with the application server initializing a server socket, a crucial communication endpoint in the system. Worker nodes, as they join the federated learning process, establish direct TCP socket connections with the server, enabling the direct sharing of crucial information like smart contract addresses. This address enables worker nodes to interact with the blockchain part of our framework, ensuring transparent and secure collaborative learning. Concurrently, workers provide metadata like their location upon enrollment.

The server then leverages geographic data for efficient cluster formation, grouping physically proximate nodes to enhance communication efficiency. This approach capitalizes on data similarities from proximity, potentially improving model accuracy. Within these clusters, direct communication is enabled through JSON files that include cluster head information, allowing seamless communication between workers. Once clusters are formed, the server shares cluster head details; worker nodes train ML models on their own data and update their models and directly transmit weights to their cluster head in a peer-to-peer manner via sockets. The cluster head performs local model weight aggregation, combining received updates from peer worker nodes to generate an updated global model. Additionally, the cluster head connects with other clusters, sharing model information and further enhancing the collaborative process. This interconnected system ensures efficient communication, cluster-based collaboration, and dynamic model aggregation for improved federated learning outcomes.

*C. Leader Selection and Contract Integration*

The process begins with the Requester deploying a smart contract onto the blockchain, requiring some tokens for deployment. Workers participate by joining the task and paying a certain amount of blockchain tokens. Subsequently, each worker retrieves their model from IPFS. This smart contract holds all the information about the workers, which is accessible to the Requester. To initiate federated learning, as discussed earlier, the Requester makes a cluster group based on worker location, and within the cluster one worker randomly gets selected to lead the aggregation process based on the information from the smart contract. All workers train their models with their own data and submit their scores to the smart contract. All workers, except the chosen leader, send their model weights to the leader for aggregation. The worker's head sends the updated weights to all workers and in IPFS. The leader then shares the updated model's hash with all other workers. Notably, the smart contract handles penalization and reward distribution based on the collected information. This process continues for subsequent rounds as well, ensuring collaborative learning and model improvement.

*D. Role of Blockchain in SDFL Process*

Blockchain technology plays a pivotal role in enhancing various aspects of the SDFL process. A smart contract committed to the blockchain efficiently coordinates the FL task, facilitating the distribution of rewards and penalization of dishonest actors. The decentralized and distributed tamper-resistant nature of blockchain ensures the reliability of the SDFL process, mitigating the risks associated with a single central authority. Its transparent and immutable ledger provides an auditable record of model updates and transactions during the SDFL process, enabling traceability and fostering accountability among participants. The cryptographic mechanisms in blockchain securely verify participants' identities, preventing unauthorized access and creating a secure environment for collaborative learning [2].

In SDFL, data privacy is upheld as raw data is not shared centrally. Instead, the model updates trained by participants at the end of each round are securely stored on the InterPlanetary File System (IPFS), safeguarding against data leakage and unauthorized access to sensitive information. By leveraging the integration of blockchain in SDFL, participants can confidently contribute to the collaborative learning process while retaining control over their private data. The combination of tamper-resistant blockchain technology and IPFS storage fortifies the security and privacy of DFL, paving the way for a robust and transparent decentralized learning ecosystem.

*E. Trust Penalization and Async Functionality*

The integration of trust penalization and asynchronous functionality augments the reliability and trustworthiness of SDFL-B.

The trust penalization mechanism addresses the presence of bad or dishonest nodes by assessing their contributions and behaviors during the learning process. Nodes are evaluated based on model updates, protocol adherence, and contribution quality, and dishonest behavior is penalized. This mechanism fosters a trustworthy and accountable learning environment, promoting fairness and encouraging honest participation. We utilize the following algorithm for trust penalization:

**Algorithm 1** Trust Penalization Algorithm

---
1. Requester (R) initializes the smart contract by depositing funds (D) to cover the task rewards.
   $$R \rightarrow SmartContractInitiation(D)$$
2. Each worker $w \in W$ who wishes to participate in the task deposits a fixed amount F of money:
   $$\forall w \in W: w \rightarrow Deposit(F)$$
   This deposit ensures commitment and creates a level playing field among workers.
3. Workers' performance is evaluated based on the evaluation score S(w):
   $$\forall w \in W: S(w)=EvaluatePerformance(w)$$
4. Bad workers with evaluation scores below the threshold T are identified:
   $$BadWorkers = \{w \in W \mid S(w) < T\}$$
   Bad workers are penalized for their suboptimal performance. Penalties are imposed on the bad workers:
   $$\forall w \in BadWorkers: Pen(w) = F \cdot P/100,$$ where P represents the penalty percentage.
5. Penalty amounts are deducted from workers' deposits, reducing their remaining deposit (D(w)) as follows:
   $$\forall w \in BadWorkers: D(w) = F - Pen(w)$$
6. The remaining deposit amount is refunded to the workers:
   $$\forall w \in W: Refund(w) = D(w)$$
   The refund process maintains equitable treatment for all participants.
7. Collected penalties are transferred back to the requester:
   $$R \leftarrow TransferPenalties(\sum_{w \in BadWorkers} Pen(w))$$
   This step ensures that penalized funds are appropriately utilized.
8. Top k workers based on specified rules are selected for reward distribution:
   $$TopKWorkers = SelectTopK(W, k)$$
   Rewards are distributed among the top workers:
   $$\forall w \in TopKWorkers: Reward(w)=R_{total}/k$$
---

This algorithm ensures a competitive and incentivized environment for workers to deliver high-quality contributions while penalizing underperforming workers. When participants know that their contributions are being evaluated and that there are consequences for dishonest behavior, they are more likely to contribute accurate and meaningful updates to the shared model through honest behavior. This promotes the convergence of the model towards a consensus that reflects the true underlying patterns in the data. The presence of penalties acts as a deterrent against malicious actions, thereby reducing the likelihood of deliberate attempts to disrupt the learning process. This, in turn, increases the reliability of model updates and the system as a whole. Participants can have more confidence in the accuracy and integrity of the shared model, leading to improved decision-making and outcomes in applications that rely on the FL process.

On the other hand, asynchronous functionality empowers nodes to contribute model updates independently, without synchronization with any other entity. This approach tackles issues related to network delays, varying computational capabilities of nodes, and potential node failures. With asynchronous updates, real-time performance is achieved, ensuring system resilience and efficiency in SDFL-B.

*1) Asynchronous Updates and Their Advantages in Distributed Collaborative Learning:* Primarily, it facilitates real-time performance enhancement, empowering individual nodes to independently update the shared model according to their pace. This dynamic approach expedites convergence, reduces training times, and frees nodes from the restrictions of synchronous communication cycles. Consequently, real-time updates enable rapid model improvements, making the system responsive to evolving data patterns. Additionally, this asynchronous feature fortifies the system's resilience against node failures, a crucial aspect in decentralized contexts like SDFL. The system's independence from the participation of every node in each communication round ensures its steadfastness amidst intermittent connectivity or node disconnections. This adaptability empowers the model to sustain progress even when certain nodes experience delays or temporary unavailability. Moreover, *the* diverse computational and communication capacities inherent in decentralized networks are effectively managed through asynchronous updates, enabling nodes to contribute as they're prepared. This strategy optimizes resource allocation, prevents any single node from becoming a bottleneck, and ensures smooth progress throughout collaborative learning. However, implementing asynchronous functionality brings about trade-offs and challenges, including:

*a) Consistency and Convergence:* Asynchronous updates introduce inconsistencies in node models, challenging convergence. Local model updates, weighted averaging, and advanced aggregation methods are needed to ensure accurate and meaningful convergence.

*b) Communication Efficiency:* Asynchronous updates reduce bottlenecks but create communication overhead due to frequent updates. Efficient communication protocols and strategies are vital to minimize latency, particularly in large networks.

*c) Addressing Stragglers:* Asynchronous systems lead to slow-updating nodes or "stragglers" that hinder training. Techniques like redundancy, adaptive learning rates, and scheduling are essential to counteract straggler effects.

*d) Balancing Blockchain Overhead:* Integrating blockchain in SDFL adds overhead from transaction verification and consensus mechanisms. Achieving an equilibrium between blockchain benefits and computational costs is pivotal.

*e) Privacy and Security Implications:* Asynchronous updates raise privacy and security concerns due to varying data exposure levels. Privacy-preserving techniques are applied during aggregation and updates to maintain data confidentiality and integrity [2].

By combining trust penalization and asynchronous functionality, the SDFL system fosters secure, efficient, and transparent collaborative learning, incentivizing the involvement of reliable nodes and enhancing the overall performance and reliability of the system.

## IV. EXPERIMENTAL SETUP

We utilized an x86_64 architecture with 16 CPU cores and 32 threads. We employed the Intel(R) Xeon(R) CPU E5-2673 v4 model, which operated within a frequency range of 1200.0000 MHz to 2300.0000 MHz. We used a dual socket, featuring 8 cores per socket, as the foundation for our experimental investigations

We assessed the performance and generalization capabilities of the Cluster-Based SDFL-B model using the well-known MNIST dataset as in our previous paper [2]. This dataset served as the foundation for assessing the model's performance and generalization capabilities. The experiment employed a Recursive architecture, inspired by the established framework for Decentralized Federated Learning with Blockchain.

This architecture, denoted as 'Net', encompasses several crucial components that contribute to the model's functionality. These components include convolutional layers (conv1 and conv2), a dropout layer (conv2_drop), and fully connected layers (fc1 and fc2). This recursive structure facilitates decentralized learning and emphasizes collaborative knowledge exchange among participating nodes.

The utilized hyperparameters were carefully selected to fine-tune the model's performance. The 'SGD' (Stochastic Gradient Descent) optimizer was chosen with a learning rate of 0.01, a momentum value of 0.5, a dampening set to 0, and no weight decay or Nesterov acceleration. These hyperparameters were tailored to strike a balance between efficient convergence and effective regularization during training. They played a pivotal role in shaping the training process and optimization strategy of the model.

Network Architecture

```
ScriptModule(
  original_name=Net
  (conv1):
   RecursiveScriptModule(original_name=Conv2d)
  (conv2):
   RecursiveScriptModule(original_name=Conv2d)
  (conv2_drop):
   RecursiveScriptModule(original_name=Dropout2d)
  (fc1): RecursiveScriptModule(original_name=Linear)
  (fc2): RecursiveScriptModule(original_name=Linear)
)
```

Model Parameter

```
{'state_dict': {'state': {},
'param_groups': [{'lr': 0.01,
'momentum': 0.5,
'dampening': 0,
'weight_decay': 0,
'nesterov': False,
'params': [0, 1, 2, 3, 4, 5, 6, 7]}]},
'name': 'SGD'}
```

The model parameters, encapsulated within the state dictionary, were initialized based on the specified hyperparameters. The model's architecture and parameters were organized into parameter groups, each containing a subset of learnable parameters. The utilization of these parameters was orchestrated by the SGD optimizer, which ensured an iterative optimization process that progressively refined the model's internal representations.

## V. RESULTS AND ANALYSIS

In our evaluations, we first analyzed scenarios involving 3 workers, both with and without the integration of blockchain technology. The results, as depicted in Fig. 2, revealed a remarkable consistency in accuracy, regardless of blockchain utilization. However, it is noteworthy that employing blockchain confers significant advantages, including enhanced trust, assurance of transparency, and a means for imposing penalties. On the flip side, when considering time efficiency, communication without blockchain emerged as the more time-effective option in the long run compared to its blockchain-enabled counterpart. This finding underscores the importance of carefully weighing the benefits and trade-offs associated with blockchain integration in decentralized systems.

To assess the scalability of our semi-decentralized federated learning framework, we conducted an in-depth analysis by calculating the average accuracy across different worker participation scenarios: 8 workers, 16 workers and 20 workers, for each epoch. As shown in Fig. 3, the outcomes reveal consistent accuracy trends across different worker counts. This observation suggests that our semi-decentralized federated learning framework exhibits promising scalability, as demonstrated by the consistent trends observed across epochs.

To ensure the reliability of the individual worker outputs, we calculated the standard deviations of accuracy for the comparison between configurations involving 8, 16 and 20 workers. Our findings, as shown in Fig. 4 indicate that the semi-decentralized federated learning framework exhibits similar consistency and reliability with 8, 16 and 20 workers. This is supported by the similar standard deviation in accuracy metrics across epochs, indicating a stable and dependable training procedure. The similar performance variability emphasizes the system's robustness and improved reliability with a greater number of workers.

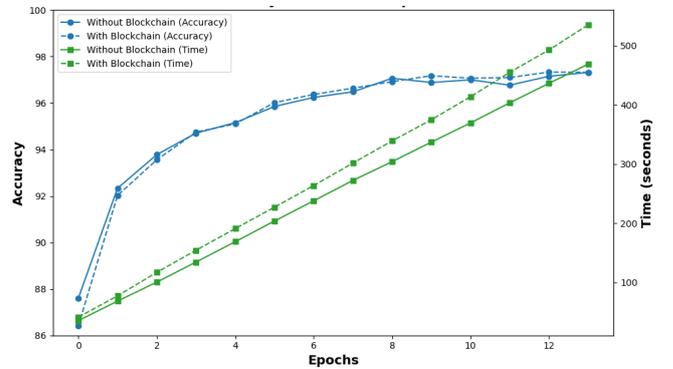

Fig. 2. Analysis of Accuracy and Time with and without Blockchain

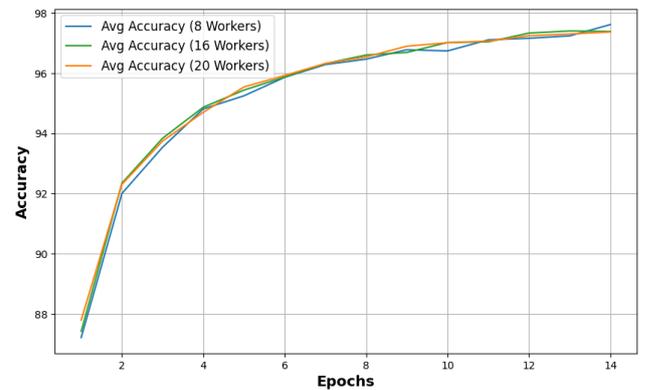

Fig. 3. Scalability Assessment (Accuracy vs Number of Epochs)

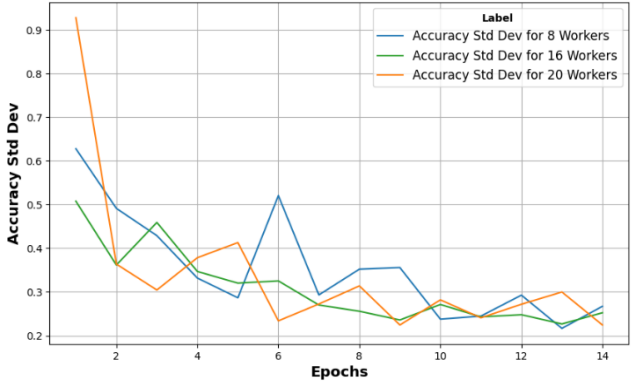

Fig. 4. Reliability Assessment (Accuracy Standard Deviations vs Number of Epochs)

The model convergence analysis involved examining the accuracy and loss curves of each worker individually as shown in Fig. 5 and Fig. 6. It was observed that, although there were slight variations in convergence rates, all workers demonstrated a clear trend of improving accuracy and diminishing loss as training progressed. This underscores the efficacy of the distributed approach in achieving a model convergence process, demonstrating its capacity to guide diverse workers and datasets toward optimal learning outcomes.

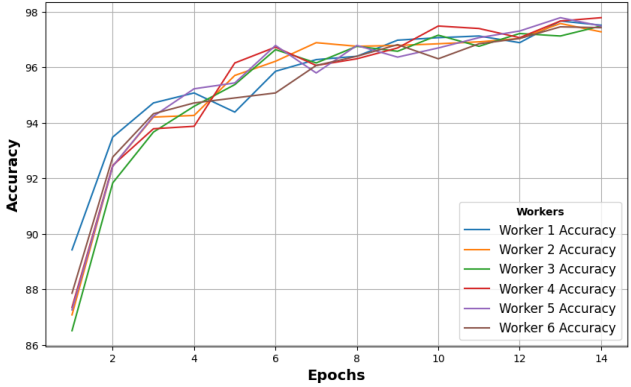

Fig. 5. Model Convergence - Accuracy Patterns

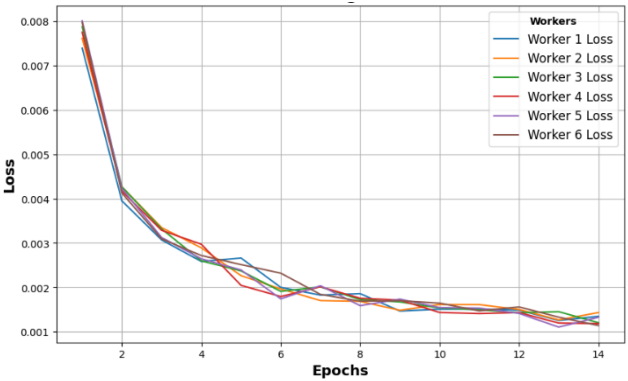

Fig. 6. Model Convergence - Loss Patterns

## VI. Discussion

### A. Impact of Trust Penalization on Node Behavior

Trust penalization can encourage workers to follow the protocol and behave honestly. Workers are incentivized to provide accurate model updates to avoid penalties that could diminish their earnings or reputation within the network. Next, the incentive for compliance is heightened, motivating workers to actively engage and fulfill their roles within the semi-decentralized federated learning process. This engenders a more engaged and cooperative group of workers. Furthermore, the phenomenon of "free-riding" may decrease as workers understand that untrustworthy actions might lead to penalties. This balanced distribution of work and participation can result in a more equitable environment.

### B. Impact of Trust Penalization on Model Performance

Trust penalization can lead to higher model performance as workers are encouraged to provide accurate and high-quality updates. Inaccurate or malicious contributions are discouraged, leading to a more accurate and reliable model over time. Furthermore, penalization can help filter out noise introduced by unreliable or intentionally malicious workers. This can prevent the model from being negatively influenced by incorrect updates, improving the convergence speed and overall quality of the model. This robustness reinforces the collaborative learning process against potential attacks or attempts to undermine it. The mechanism also ensures fairness among workers by holding everyone accountable for their contributions. This can lead to a more balanced distribution of rewards and prevent a single malicious worker from disproportionately affecting the model.

### C. Asynchronous Functionality in SDFL-B Across Varied Scenarios

Asynchronous updates mean that workers can train their models independently and update them whenever they are ready. This reduces the need for strict synchronization among workers, which can lead to more efficient resource utilization and faster convergence. In scenarios with many workers, synchronous updates might lead to bottlenecks and delays due to the need for coordinated updates. Asynchronous updates can mitigate this issue. Furthermore, asynchronous updates can enhance fault tolerance. If one or more workers experience connectivity issues or temporary failures, it won't disrupt the entire training process since other workers can continue to make progress.

### D. Generic Codebase Paradigm

Our implementation is highly adaptable and modular. Increasing the number of clusters and workers in clusters only requires a simple change in parameters when the requester is initializing the network topology. Likewise, modifying the ML model and hyperparameters is also handled by the requester during the initialization phase. Our codebase seamlessly supports several popular ML Python libraries, including Pytorch, TensorFlow and ScikitLearn. This modular implementation allows different application scenarios to easily leverage our code with minimal configurations. This standardization and modularity significantly decrease redundant implementations of SDFL-B, fostering enhanced adaptability within varied contexts.

### E. Limitations and Potential Areas For Future Research

In asynchronous settings, there's a challenge related to selecting leaders. The complexity emerges because leaders chosen at random might be bad workers and affect the performance of the model by pushing the bad weights to the IPFS which can impact the entire network. Further research in this direction is needed to optimize this aspect of the system.

## VII. CONCLUSION

In conclusion, this research introduces an innovative approach that addresses the challenges of scalability and reliability in Decentralized Federated Learning through the integration of blockchain technology. By combining semi-decentralized federated learning with blockchain, the proposed system establishes a fair, secure, and transparent environment for collaborative machine learning while preserving data privacy. The incorporation of a trust penalization mechanism enhances the trustworthiness of participating nodes, fostering reliability and accountability, while asynchronous functionality ensures efficient and robust model updates. The results from experimental evaluations demonstrate the efficacy of the approach in promoting scalability and trustworthiness in SDFL-B systems. The discussed impact of trust penalization on node behavior and model performance underscores the positive influence of this mechanism. Asynchronous functionality's adaptability across various scenarios highlights its efficiency and fault tolerance benefits. Additionally, conducting user studies and experimental evaluations to validate the trust penalization algorithm's correctness and efficacy would be an essential avenue for future work. The insights gained from this study, coupled with further research, could focus on refining leader selection in asynchronous settings and establishing effective communication protocols for updating leaders' models before aggregation. This research contributes to the advancement and standardization of decentralized machine learning methodologies, enabling the development of scalable and reliable SDFL-B systems that cater to diverse applications across domains.

## REFERENCES

[1] K. Bonawitz *et al.*, "Towards federated learning at scale: system design," in *Machine Learning and Systems 1 (MLSys 2019)*, 2019, pp. 374–388.

[2] A. Jaberzadeh, A. K. Shrestha, F. A. Khan, M. A. Shaikh, B. Dave, and J. Geng, "Blockchain-based federated learning: incentivizing data sharing and penalizing dishonest behavior," Blockchain and Applications, 5th International Congress, BLOCKCHAIN 2023, Lecture Notes in Networks and Systems, Springer, Cham, 2023., in press.

[3] E. T. M. Beltrán *et al.*, "Decentralized Federated Learning: Fundamentals, state of the art, frameworks, trends, and challenges," Nov. 2022, Accessed: Aug. 21, 2023. [Online]. Available: http://arxiv.org/abs/2211.08413.

[4] K. Zhang, X. Song, C. Zhang, and S. Yu, "Challenges and future directions of secure federated learning: a survey," *Front. Comput. Sci.*, vol. 16, no. 5, Dec. 2022, doi: 10.1007/s11704-021-0598-z.

[5] K. Liu, Z. Yan, X. Liang, R. Kantola, and C. Hu, "A survey on blockchain-enabled federated learning and its prospects with digital twin," *Digit. Commun. Networks*, Aug. 2022, doi: 10.1016/j.dcan.2022.08.001.

[6] A. K. Shrestha and J. Vassileva, "User Data Sharing Frameworks: A Blockchain-Based Incentive Solution," in *2019 IEEE 10th Annual Information Technology, Electronics and Mobile Communication Conference (IEMCON)*, Oct. 2019, pp. 0360–0366, doi: 10.1109/IEMCON.2019.8936137.

[7] L. He, A. Bian, and M. Jaggi, "COLA: Decentralized linear learning," *Adv. Neural Inf. Process. Syst.*, vol. 31, 2018.

[8] S. Augenstein, A. Hard, K. Partridge, and R. Mathews, "Jointly learning from decentralized (federated) and centralized data to mitigate distribution shift," NeurIPS 2021 Workshop, Dec. 2021.

[9] T. Wang, Y. Liu, X. Zheng, H. N. Dai, W. Jia, and M. Xie, "Edge-based communication optimization for distributed federated learning," *IEEE Trans. Netw. Sci. Eng.*, vol. 9, no. 4, pp. 2015–2024, 2022, doi: 10.1109/TNSE.2021.3083263.

[10] S. Savazzi, M. Nicoli, M. Bennis, S. Kianoush, and L. Barbieri, "Opportunities of federated learning in connected, cooperative, and automated industrial systems," *IEEE Commun. Mag.*, vol. 59, no. 2, pp. 16–21, Feb. 2021, doi: 10.1109/MCOM.001.2000200.

[11] A. Bellet, A. M. Kermarrec, and E. Lavoie, "D-Cliques: Compensating for data heterogeneity with topology in decentralized federated learning," *Proc. IEEE Symp. Reliab. Distrib. Syst.*, Sept 2022, pp. 1–11, doi: 10.1109/SRDS55811.2022.00011.

[12] E. Georgatos, C. Mavrokefalidis, and K. Berberidis, "Fully distributed federated learning with efficient local cooperations," in *ICASSP 2023 - 2023 IEEE International Conference on Acoustics, Speech and Signal Processing (ICASSP)*, Jun. 2023, pp. 1–5, doi: 10.1109/ICASSP49357.2023.10095741.

[13] Y. Qu, C. Xu, L. Gao, Y. Xiang, and S. Yu, "FL-SEC: Privacy-preserving decentralized federated learning using SignSGD for the internet of artificially intelligent things," *IEEE Internet Things Mag.*, vol. 5, no. 1, pp. 85–90, May 2022, doi: 10.1109/IOTM.001.2100173.

[14] M. Yemini, R. Saha, E. Ozfatura, D. Gunduz, and A. J. Goldsmith, "Semi-decentralized federated learning with collaborative relaying," in *IEEE International Symposium on Information Theory - Proceedings*, Jun. 2022, pp. 1471–1476, doi: 10.1109/ISIT50566.2022.9834707.

[15] M. S. Al-Abiad, M. Obeed, M. J. Hossain, and A. Chaaban, "Decentralized aggregation for energy-efficient federated learning via d2d communications," *IEEE Trans. Commun.*, vol. 71, no. 6, pp. 3333–3351, Jun. 2023, doi: 10.1109/TCOMM.2023.3253718.

[16] F. P. C. Lin, S. Hosseinalipour, S. S. Azam, C. G. Brinton, and N. Michelusi, "Semi-decentralized federated learning with cooperative d2d local model aggregations," *IEEE J. Sel. Areas Commun.*, vol. 39, no. 12, pp. 3851–3869, Dec. 2021, doi: 10.1109/JSAC.2021.3118344.

[17] R. Wang and W. T. Tsai, "Asynchronous federated learning system based on permissioned blockchains," *Sensors (Basel).*, vol. 22, no. 4, Feb. 2022, doi: 10.3390/S22041672.